\documentclass[11pt]{article}
\usepackage[final]{acl}

\usepackage{times}
\usepackage{latexsym}
\usepackage[T1]{fontenc}
\usepackage[utf8]{inputenc}
\usepackage{microtype}
\usepackage{graphicx}
\usepackage{fixltx2e}
\usepackage{booktabs}
\usepackage{amsmath}
\usepackage{amssymb}
\usepackage{array}
\usepackage{multirow}
\usepackage{xcolor}
\usepackage{float}
\usepackage{placeins}
\usepackage{tikz}
\usepackage{pgfplots}
\pgfplotsset{compat=1.18}
\usepackage{tikz-cd}
\usetikzlibrary{shapes.geometric, arrows.meta, positioning, fit, backgrounds, calc}
\title{L3Cube-MahaPOS: A Marathi Part-of-Speech Tagging Dataset and BERT Models}

\author{
Hariom Ingle$^{1,3}$ \quad Ronit Ghode$^{1,3}$ \quad Ishwari Gondkar$^{1,3}$ \quad Jidnyasa Harad$^{1,3}$ \quad Raviraj Joshi$^{2,3}$ \\[4pt]
$^{1}$Department of Information Technology, PICT, Pune, India \\
$^{2}$Indian Institute of Technology Madras, Chennai, India \\
$^{3}$L3 Cube Labs, Pune, India \\
\small\texttt{\url{ravirajoshi@gmail.com}}
}

\begin{document}
\maketitle

\begin{abstract}
Part-of-Speech (POS) tagging is a foundational NLP task underpinning machine translation, information extraction, and syntactic parsing. Despite Marathi being spoken by over 83 million people and ranking among the top twenty most spoken languages worldwide, it remains severely under-resourced in annotated corpora and standardised evaluation benchmarks. Marathi presents unique challenges for computational modelling owing to its rich morphology, relatively free word order, lack of capitalisation conventions, and pervasive code-mixing with Hindi and English.

We introduce \textbf{L3Cube-MahaPOS}, a gold-standard POS tagging dataset for Marathi comprising 32,354 manually annotated sentences drawn from news text. Annotation was performed entirely manually by a team of Marathi-proficient annotators following a 16-tag Universal Dependencies-aligned scheme. A structured preprocessing pipeline covering Unicode normalisation, Devanagari-aware tokenisation, and noise filtering ensures label consistency across all splits.

We benchmark the dataset across six model families spanning HMM, CRF, BiLSTM, BiLSTM+CharCNN, MuRIL, and the Marathi-specific transformer \textit{MahaBERT-v2}. The best system achieves 88.67\% token-level accuracy and a macro-F1 of \textbf{81.67\%} over 15 evaluated tag classes. We release the dataset, annotation guidelines, and trained model checkpoints to foster further research in Marathi NLP.
\end{abstract}

\section{Introduction}
\label{sec:intro}

Natural Language Processing has matured into a central pillar of artificial intelligence, transforming how computational systems comprehend and produce human language. Among the elemental tasks within this field, Part-of-Speech (POS) tagging occupies a privileged position: by assigning a grammatical category---noun, verb, adjective, adverb, and so forth---to every token in a sentence, POS tagging supplies downstream components with the structural metadata they require to function reliably. Syntactic parsers, named entity recognisers, machine translation decoders, and question-answering systems all trace a portion of their performance directly to the quality of the POS labels they consume.

For high-resource languages such as English, French, and Mandarin, the combination of large treebanks, shared tasks, and pre-trained language models has pushed tagging accuracy above 97\% on standard benchmarks~\citep{manning2011part}. Tools such as Stanford NLP, NLTK, and spaCy provide production-ready taggers that require minimal configuration. The landscape for Indian languages, however, is strikingly different. Marathi is the official language of Maharashtra, India, ranking among the top twenty most spoken languages worldwide; yet within computational linguistics, it remains firmly categorised as a low-resource language~\citep{kulkarni2021l3cubemahacorpus,joshi2022l3cube_mahacorpus,joshi2022l3cube}.

The NLP challenges specific to Marathi are both numerous and interacting. First, Marathi is morphologically rich: a single verb stem may take dozens of inflected forms encoding tense, aspect, mood, number, gender, and honorific level simultaneously. This inflectional density multiplies vocabulary size and increases lexical sparsity. Second, the language exhibits relatively free word order, meaning syntactic roles must be inferred from morphological cues rather than positional patterns---the dominant strategy for English-centric models. Third, Marathi lacks capitalisation conventions that English-based systems exploit as features for proper noun detection. Fourth, code-mixing with English and Hindi is pervasive in contemporary written Marathi, particularly in digital and journalistic text. Finally, dialectal variation across geographic regions introduces orthographic inconsistencies that frustrate standard tokenisers.

Prior computational work on Marathi has concentrated on tasks with immediate societal applications. \citet{kulkarni2021l3cubemahacorpus} introduced the L3Cube-MahaSent benchmark for sentiment analysis; \citet{litake2022mahaner} presented L3Cube-MahaNER for named entity recognition; \citet{pingle2023mahasentmd} extended sentiment analysis to a multi-domain setting; and \citet{chaudhari2024mahsocialner} introduced social media NER for Marathi. POS tagging, however, has received comparatively little attention, and existing resources suffer from small size, non-standard tag sets, absent inter-annotator reliability reports, or restricted access.

This paper addresses that gap through three coordinated contributions. First, we describe the construction of \textbf{L3Cube-MahaPOS}, a carefully curated and manually annotated Marathi POS corpus comprising 32,354 sentences from real-world news text, annotated with 16 UD-aligned tags and subjected to a rigorous quality-assurance protocol. Second, we define a reproducible preprocessing and annotation pipeline---normalisation, tokenisation, ambiguity adjudication, and split stratification---that future corpus builders can adapt. Third, we present a comprehensive benchmarking study evaluating six model families on the corpus, ranging from classical statistical models to the state-of-the-art Marathi BERT variant~\citep{joshi2022l3cube_mahacorpus}, providing the community with a transparent, reproducible baseline.

The contributions of this paper may be summarised as follows:
\begin{itemize}
    \item We release \textbf{L3Cube-MahaPOS}\footnote{\scriptsize\href{https://huggingface.co/datasets/l3cube-pune/marathi-pos-tagger}{MahaPOS Dataset} (l3cube-pune/marathi-pos-tagger)}, one of the first large-scale, manually annotated POS tagging datasets for Marathi, along with the corresponding fine-tuned \textbf{MahaPOS-BERT}\footnote{\scriptsize\href{https://huggingface.co/l3cube-pune/marathi-pos-tagger}{MahaPOS-BERT Model} (l3cube-pune/marathi-pos-tagger)}\footnote{\scriptsize\url{https://github.com/l3cube-pune/MarathiNLP}} model, with a well-documented annotation scheme.
    \item We adopt a 16-tag set aligned with the Universal Dependencies framework, ensuring interoperability with multilingual resources.
    \item We introduce a structured data preparation pipeline covering normalisation, tokenisation, and conflict resolution.
    \item We benchmark six model families---from HMM to transformer-based architectures---establishing strong, reproducible baselines.
    \item We provide a detailed error analysis that reveals the principal sources of tagging difficulty in Marathi, guiding future modelling efforts.
\end{itemize}

The remainder of this paper is organised as follows. Section~\ref{sec:related} surveys relevant prior work. Section~\ref{sec:dataset} details corpus construction. Section~\ref{sec:experiments} describes experimental methodology. Section~\ref{sec:results} presents and analyses results. Section~\ref{sec:limitations} acknowledges limitations, and Section~\ref{sec:conclusion} concludes.

\section{Related Work}
\label{sec:related}

\subsection{POS Tagging: A Longitudinal Perspective}

Early POS taggers operated through hand-crafted rule systems that encoded morphological and syntactic knowledge directly as deterministic transformations~\citep{brill1992simple}. Although interpretable and language-agnostic in principle, such systems demanded prohibitive expert effort and generalised poorly across domains. The subsequent emergence of probabilistic sequence models---notably Hidden Markov Models and Maximum Entropy classifiers---shifted the paradigm toward corpus-driven learning, enabling taggers to acquire contextual patterns automatically from annotated data. \citet{ratnaparkhi1996maximum} first demonstrated that a maximum entropy framework could combine diverse, overlapping contextual features without the independence assumptions imposed by Hidden Markov Models, achieving accuracy competitive with the best contemporary taggers on the Penn Treebank~\citep{marcus1993building}. \citet{brants2000tnt} subsequently showed, with the TnT tagger, that a carefully smoothed second-order Markov model could match or exceed maximum-entropy approaches while remaining considerably faster to train, a finding that renewed interest in the practical trade-offs between generative and discriminative formulations. Conditional Random Fields~\citep{lafferty2001conditional} further improved upon their predecessors by relaxing the independence assumptions of HMMs, allowing richer feature templates that capture arbitrary overlapping evidence from both the past and the future of an observation sequence. \citet{toutanova2003feature} pushed discriminative tagging further still with a cyclic dependency network that conditions jointly on left and right tag contexts, reaching 97.24\% accuracy on the Penn Treebank WSJ and establishing what remained, for over a decade, the benchmark feature-engineered English tagger.

The deep learning era introduced representation learning as an alternative to manual feature engineering. Recurrent architectures, particularly Long Short-Term Memory networks~\citep{hochreiter1997long} and their bidirectional extensions, learn dense token representations from context implicitly, bypassing the need for morphological analysers or gazetteers. Convolutional character-level encoders added robustness to unseen words~\citep{ma2016end}. The introduction of attention-based transformer models~\citep{vaswani2017attention}---and their fine-tunable pre-trained instantiations such as BERT~\citep{devlin2019bert}---subsequently established new state-of-the-art results across almost every sequence labelling task, including POS tagging on the Penn Treebank and Universal Dependencies benchmarks.

\subsection{POS Tagging for Indian Languages}

Indian languages collectively present a challenging frontier for sequence labelling owing to their Dravidian or Indo-Aryan morphological complexity, script diversity, and limited annotation resources. Dedicated research programmes have produced annotated corpora for Hindi, Bengali, Tamil, and Telugu, yet coverage remains uneven and quality varies considerably across resources.

For Hindi, efforts culminating in the Hindi Dependency Treebank~\citep{palmer2009hindi} provided a foundation for syntactic and POS studies. \citet{singh2006morphological} demonstrated that morphologically motivated features---chiefly suffix and inflectional cues---could offset the comparative scarcity of annotated Hindi data, reporting tagging accuracy competitive with resource-rich English taggers despite a training corpus several orders of magnitude smaller; subsequent work leveraged multilingual BERT variants to achieve competitive accuracy despite limited data. Bengali POS tagging benefitted from the NLTR corpus and from pre-trained language models such as BanglaBERT. Tamil and Telugu have seen similar, though smaller-scale, dataset creation initiatives. A cross-cutting observation from these efforts is that performance tends to degrade sharply for morphologically rare tag classes and for out-of-vocabulary tokens produced by productive derivational processes---a finding that our Marathi results corroborate.

\subsection{POS Tagging Beyond English: A Cross-Linguistic View}

Outside the Indian subcontinent, POS tagging research has grappled with an equally diverse set of script- and morphology-specific obstacles, and the resulting body of work offers useful points of comparison for Marathi. For Arabic, whose templatic morphology and absence of short-vowel diacritics in everyday text create severe lexical ambiguity, \citet{diab2004automatic} introduced an SVM-based pipeline that jointly tokenises clitics, assigns POS tags, and chunks base phrases, reporting accuracy above 95\% despite the lack of explicit word boundaries for many bound morphemes. Chinese poses a distinct difficulty, since word segmentation and POS assignment are intertwined in the absence of whitespace; \citet{nglow2004chinese} compared word-based and character-based tagging strategies combined with maximum-entropy classification and found that the optimal strategy depends on whether segmentation is performed jointly with or prior to tagging. Japanese shares this segmentation-tagging interdependence, and \citet{kudo2004applying} showed that Conditional Random Fields could be adapted to settings where word boundaries are themselves ambiguous, improving over earlier HMM- and MEMM-based Japanese morphological analysers by reducing label- and length-bias effects. For Korean, an agglutinative language with productive morpheme-internal alternation, \citet{lee2002syllable} addressed the dominant source of tagging error---unknown morphemes---through a syllable-pattern-based estimation method combined with a rule-based hybrid tagger. Taken together, these studies reinforce a recurring theme also visible in our Marathi results: languages that lack reliable orthographic cues (capitalisation, whitespace, or diacritics) shift the tagging burden onto morphological and contextual features, and the most effective systems are those that explicitly model this shift rather than treating the language as a script-transposed variant of English. Cross-lingual treebank initiatives such as Universal Dependencies~\citep{nivre2016universal} have since sought to harmonise annotation schemes across this typological diversity, a framework we adopt for L3Cube-MahaPOS.

\subsection{Marathi NLP}

Computational resources for Marathi have expanded noticeably over the past half-decade. The L3Cube-MahaSent dataset~\citep{kulkarni2021l3cubemahacorpus} provided a foundational benchmark for Marathi sentiment analysis. \citet{litake2022mahaner} presented L3Cube-MahaNER, the first gold-standard NER dataset for Marathi, benchmarked across CNN, LSTM, and transformer models with MahaBERT achieving the best results. \citet{pingle2023mahasentmd} extended this to L3Cube-MahaSent-MD, a multi-domain sentiment dataset covering movie reviews, tweets, and subtitles. \citet{chaudhari2024mahsocialner} introduced L3Cube-MahaSocialNER, targeting informal social media text for NER. More recently, \citet{kowtal2025mahaemotions} presented L3Cube-MahaEmotions for fine-grained emotion recognition in Marathi using LLM-based annotation. The \texttt{MahaBERT-v2} model~\citep{joshi2022l3cube_mahacorpus}, pre-trained on a large Marathi text collection, has become the de facto starting point for Marathi downstream tasks and is the strongest model in our benchmarking study.

Prior POS tagging work for Marathi is sparse. Existing studies have typically employed rule-based or small-scale statistical approaches with tag sets that lack standardisation, making cross-study comparison impossible. None of the publicly known datasets reach a size or annotation quality that would support rigorous neural benchmarking. The Universal Dependencies project includes a small Marathi treebank (UD\_Marathi-UFAL), but its scale---approximately 466 sentences---is insufficient for reliable fine-tuning of large language models. The present paper directly addresses these gaps.

\subsection{Multilingual and Transfer Learning Approaches}

Recent advances in multilingual pre-training---including mBERT~\citep{devlin2019bert}, XLM-R~\citep{conneau2020unsupervised}, and MuRIL---have enabled cross-lingual transfer for low-resource languages, including Marathi. These models, pre-trained on Wikipedia dumps across 100+ languages, provide a useful baseline but typically underperform language-specific models trained on larger, in-domain corpora. IndicBERT and related Indic-family models represent an intermediate point, leveraging shared sub-word vocabularies across related Indian languages. Our benchmarking study includes comparisons that illuminate the trade-offs among these strategies.

\section{Dataset Construction}
\label{sec:dataset}

\subsection{Data Sources and Collection}

The raw corpus for \textbf{L3Cube-MahaPOS} was assembled from Marathi news portals covering diverse topical domains: politics, sports, culture, technology, and local affairs. News text was selected for three reasons: (i) it exhibits formal register with relatively consistent orthography, facilitating reproducible preprocessing; (ii) it provides broad vocabulary coverage including technical and domain-specific terminology; and (iii) it captures contemporary Marathi as actively used in published media rather than archaic literary forms.

HTML content was extracted using domain-specific scrapers, after which boilerplate elements---navigation menus, advertisements, metadata blocks---were removed via heuristic filtering. Duplicate and near-duplicate articles were identified using MinHash locality-sensitive hashing and discarded. The resulting plain-text collection underwent sentence boundary detection using a rule-based system tuned to Marathi punctuation conventions (the Devanagari danda, a vertical bar character, as the primary sentence terminator, augmented by full stops in mixed-script contexts).

\subsection{Preprocessing Pipeline}

\begin{figure}[t]
\centering
\begin{tikzpicture}[
    node distance=0.55cm,
    box/.style={
        rectangle,
        rounded corners=4pt,
        draw=black!70,
        fill=blue!10,
        text width=3.2cm,
        minimum height=0.9cm,
        align=center,
        font=\small
    },
    arrow/.style={-Stealth, thick, draw=gray!80}
]
\node[box] (norm)   {1.~Unicode\\Normalisation};
\node[box, below=of norm]   (tok)    {2.~Devanagari\\Tokenisation};
\node[box, below=of tok]    (filter) {3.~Noise\\Filtering};
\node[box, below=of filter] (annot)  {4.~POS\\Annotation};
\draw[arrow] (norm)   -- (tok);
\draw[arrow] (tok)    -- (filter);
\draw[arrow] (filter) -- (annot);
\end{tikzpicture}
\caption{Four-stage preprocessing pipeline for L3Cube-MahaPOS corpus construction.}
\label{fig:pipeline}
\end{figure}
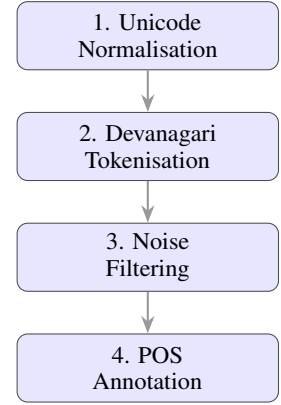

A four-stage preprocessing pipeline was applied uniformly across all data splits, as illustrated in Figure~\ref{fig:pipeline}.

\FloatBarrier

\paragraph{Stage 1 — Unicode Normalisation.}
All text was converted to Unicode NFC form. Zwj/Zwnj characters appearing at morpheme boundaries---common artefacts of digital Marathi text---were retained where they alter character appearance but removed where they are semantically redundant. Visually identical but code-point-distinct Devanagari characters (e.g., \textit{anusvara} variants) were canonicalised to a single form.

\paragraph{Stage 2 — Tokenisation.}
A Devanagari-aware tokeniser split text on whitespace and explicit punctuation boundaries. Compound postpositions and clitics attached to nominal stems without whitespace (a frequent pattern in Marathi) were handled through a small exception lexicon developed by native-speaker annotators. English words embedded in Marathi sentences were treated as single tokens and assigned appropriate UD tags.

\paragraph{Stage 3 — Noise Filtering.}
Tokens consisting exclusively of numerals with currency symbols, URL fragments, email addresses, emoticons, or HTML residues were either normalised to placeholder tokens (\texttt{<NUM>}, \texttt{<URL>}) or removed from the sentence, depending on whether their removal altered syntactic coherence. Sentences shorter than three tokens or longer than 120 tokens were excluded.

\paragraph{Stage 4 — POS Annotation.}
The cleaned token sequences were passed to the annotation workflow described in Section~\ref{sec:annotation}.

\subsection{Annotation Scheme}

We adopt a 16-category tag set aligned with the Universal Dependencies v2 framework, extended with a Marathi-specific postposition tag (\textsc{postp}) that captures a grammatical class not cleanly distinguished within the standard UD \textsc{adp} category in Marathi. Table~\ref{tab:tagset} provides the complete tag inventory with counts from the full corpus. While all 16 tags are present in the annotated data and used during model training, \textsc{postp} is excluded from primary evaluation metrics owing to its extremely low corpus frequency (23 tokens total; 3 in the test set); the full rationale is provided in Section~\ref{sec:experiments}.

\begin{table*}[t]
\centering
\small
\setlength{\tabcolsep}{10pt}
\begin{tabular}{llrc}
\toprule
\textbf{Tag} & \textbf{Description} & \textbf{Count} & \textbf{Eval} \\
\midrule
NOUN  & Common noun              & 144,597 & \checkmark \\
VERB  & Main verb                &  58,837 & \checkmark \\
PUNCT & Punctuation              &  50,044 & \checkmark \\
ADP   & Adposition               &  39,897 & \checkmark \\
ADJ   & Adjective                &  35,527 & \checkmark \\
ADV   & Adverb                   &  28,127 & \checkmark \\
AUX   & Auxiliary verb           &  23,155 & \checkmark \\
NUM   & Numeral                  &  15,711 & \checkmark \\
DET   & Determiner               &  14,145 & \checkmark \\
PRON  & Pronoun                  &  14,091 & \checkmark \\
PROPN & Proper noun              &  10,126 & \checkmark \\
CCONJ & Coordinating conjunction &   9,600 & \checkmark \\
PART  & Particle                 &   4,591 & \checkmark \\
SCONJ & Subordinating conjunction &  3,254 & \checkmark \\
INTJ  & Interjection             &   1,440 & \checkmark \\
POSTP & Postposition             &      23 & ---        \\
\midrule
\multicolumn{2}{l}{\textbf{Total tokens}} & \textbf{453,165} & \\
\bottomrule
\end{tabular}
\caption{Full tag set and token-level distribution across all splits of L3Cube-MahaPOS. \textbf{Eval} (\checkmark) marks tags included in primary evaluation. \textsc{postp} (---) is present in training but excluded from reported macro-F1 due to critically low test support ($n = 3$).}
\label{tab:tagset}
\end{table*}

The \texttt{X} tag (foreign or otherwise unclassifiable tokens) was used during annotation but subsequently removed from model training: any sentence containing at least one \texttt{X}-tagged token was excluded entirely from all data splits, preserving label coherence.

\subsection{Annotation Procedure}
\label{sec:annotation}

Annotation was performed entirely manually by a team of Marathi-proficient annotators. The full corpus of 32,354 sentences was divided into roughly equal portions and distributed across team members, each of whom tagged their assigned sentences independently. Before tagging began, all annotators were familiarised with a common set of written guidelines adapted from the UD annotation manual, supplemented with Marathi-specific decision trees covering postpositional clitics, verbal compounds, and code-mixed tokens.

To assess annotation consistency, a subset of sentences was independently tagged by multiple team members. The primary sources of disagreement were \textsc{adj}/\textsc{adv} ambiguity in participial constructions and \textsc{noun}/\textsc{propn} boundaries for institutionalised proper nouns.

Disputed and ambiguous labels were resolved through group discussion among the annotators, with the majority label adopted. Decisions were recorded in an amendment log that was used to update the shared guidelines iteratively, improving consistency across later batches. A final validation pass flagged any token whose label conflicted with the majority label assigned to the same surface form in unambiguous contexts across the corpus.

\subsection{Dataset Splits}

The final L3Cube-MahaPOS corpus comprises 32,354 sentences across training (22,652), validation (4,848), and test (4,854) splits, stratified to maintain proportional representation of all 16 tag classes and balanced domain coverage. Table~\ref{tab:splits} summarises key statistics.

\begin{table}[t]
\centering
\small
\begin{tabular}{lrrr}
\toprule
\textbf{Split} & \textbf{Sentences} & \textbf{Tokens} & \textbf{Avg.\ Len.} \\
\midrule
Train      & 22,652 & 332,418 & 14.7 \\
Validation &  4,848 &  71,163 & 14.7 \\
Test       &  4,854 &  68,878 & 14.2 \\
\midrule
Total      & 32,354 & 472,459 & 14.6 \\
\bottomrule
\end{tabular}
\caption{L3Cube-MahaPOS split statistics. Avg.\ Len.\ is mean sentence length in tokens.}
\label{tab:splits}
\end{table}

\section{Experimental Setup}
\label{sec:experiments}

\subsection{Models}

We evaluate six model configurations representing three methodological generations:

\paragraph{Classical Sequence Models.}
\textit{Hidden Markov Model (HMM):} A first-order HMM with add-one smoothing for unseen emissions. Despite its simplicity, it provides a theoretically grounded lower bound. \textit{Conditional Random Field (CRF):} A linear-chain CRF with a rich feature template comprising unigram and bigram surface forms, character suffix features of lengths 1--4, prefix features of lengths 1--3, and a binary indicator for digit-containing tokens. L-BFGS optimisation was used with L2 regularisation ($\lambda = 0.1$).

\paragraph{Neural Recurrent Models.}
\textit{BiLSTM:} A two-layer Bidirectional LSTM operating over 100-dimensional word embeddings trained from scratch, with a CRF decoding head (BiLSTM-CRF). Dropout of 0.5 is applied after each LSTM layer. \textit{BiLSTM + Character CNN:} The above architecture augmented with character-level convolutional embeddings (filter sizes 3, 4, 5; 30 filters each) concatenated to the word embeddings before the LSTM.

\paragraph{Transformer-Based Models.}
\textit{L3Cube-MahaPOS-BERT (MahaBERT-v2):} The \texttt{l3cube-pune/marathi-bert-v2} model~\citep{joshi2022l3cube_mahacorpus} fine-tuned for token classification. This model was pre-trained on a large Marathi corpus and encodes subword representations via a WordPiece vocabulary. A linear classification head over the final hidden state of the first subword of each token is used for prediction. \textit{MuRIL:} Google's Multilingual Representations for Indian Languages, included as a cross-lingual transfer baseline.

\subsection{Training Details}

All neural models were trained on an NVIDIA Tesla T4 GPU (15.6 GB). For L3Cube-MahaPOS-BERT, we used the following hyperparameters determined by a grid search over validation macro-F1: learning rate $5 \times 10^{-5}$, batch size 16, maximum sequence length 512, cosine learning rate schedule with 10\% warmup, weight decay 0.01, and a maximum of 10 training epochs with early stopping at patience 3. The model converged at epoch 3 (best validation macro-F1) and was restored from that checkpoint for final test evaluation.

Subword-to-token label alignment follows the ``first subword'' strategy: the label of a word is assigned to its first subword token; subsequent subword continuations receive the ignore label $-100$ and are excluded from loss computation and metric calculation.

\subsection{Evaluation Protocol}

Performance is reported using four metrics: \textit{token-level accuracy}, \textit{weighted macro precision}, \textit{weighted macro recall}, and \textit{macro-F1} (unweighted across all classes). All primary results are reported over \textbf{15 tags}, with \textsc{postp} excluded from evaluation. This decision is justified on statistical grounds: \textsc{postp} has a test-set support of only 3 tokens (0.004\% of the test set), making its per-class F1 estimate statistically meaningless---any single misclassification shifts its F1 by over 33 percentage points. Including such an under-represented class in the macro average introduces noise rather than signal and would misrepresent the true tagging capability of the system across well-attested grammatical categories.

For completeness and transparency, the full 16-tag macro-F1 (including \textsc{postp}) is also reported in Table~\ref{tab:overall} as a secondary figure.

\section{Results and Analysis}
\label{sec:results}

\subsection{Overall Performance}

Table~\ref{tab:overall} compares all model families on the test set. All macro-F1 figures are computed over 15 tags (\textsc{postp} excluded). L3Cube-MahaPOS-BERT achieves the highest scores on all metrics, confirming the value of large-scale Marathi-specific pre-training for this task.

\begin{table}[t]
\centering
\small
\setlength{\tabcolsep}{4pt}
\begin{tabular}{lcccc}
\toprule
\textbf{Model} & \textbf{Acc.} & \textbf{P} & \textbf{R} & \textbf{F1$_\text{mac}$} \\
\midrule
HMM               & 72.1  & 63.4  & 61.8  & 63.1 \\
CRF               & 80.6  & 72.9  & 71.5  & 72.8 \\
BiLSTM-CRF        & 84.3  & 76.8  & 75.4  & 76.6 \\
BiLSTM+CharCNN    & 85.7  & 78.2  & 77.1  & 78.3 \\
MuRIL             & 86.9  & 79.4  & 78.8  & 79.7 \\
\textbf{MahaPOS-BERT}$^\dagger$   & \textbf{88.67} & \textbf{88.63} & \textbf{76.49} & \textbf{81.67} \\
\bottomrule
\end{tabular}
\caption{Test-set results (\%) on L3Cube-MahaPOS. Macro-F1 over 15 tags (\textsc{postp} excluded); see \S\ref{sec:experiments}. $^\dagger$MahaPOS-BERT = \texttt{marathi-bert-v2} fine-tuned on L3Cube-MahaPOS; 16-tag F1 = 76.57\% (secondary reference).}
\label{tab:overall}
\end{table}

\subsection{Per-Tag Analysis}

Table~\ref{tab:pertag} details per-class precision, recall, and F1 for L3Cube-MahaPOS-BERT across the 15 evaluated tags. High-frequency tags (\textsc{noun}, \textsc{verb}, \textsc{punct}) achieve F1 scores above 91\%, reflecting adequate training signal. Tags with moderate support but morphological overlap---\textsc{adj}, \textsc{adv}, \textsc{part}---cluster around 75\%, and \textsc{propn} and \textsc{intj} represent the principal sources of macro-F1 depression owing to class imbalance and surface-form ambiguity, respectively.

\begin{table*}[t]
\centering
\small
\setlength{\tabcolsep}{8pt}
\begin{tabular}{lrrrr}
\toprule
\textbf{Tag} & \textbf{Precision} & \textbf{Recall} & \textbf{F1-score} & \textbf{Support} \\
\midrule
PUNCT  & 96.96 & 97.96 & 97.45 & 7{,}679 \\
AUX    & 94.73 & 95.97 & 95.35 & 3{,}503 \\
VERB   & 93.65 & 92.43 & 93.04 & 8{,}816 \\
ADP    & 91.87 & 93.57 & 92.71 & 6{,}051 \\
NOUN   & 91.30 & 91.92 & 91.61 & 21{,}910 \\
NUM    & 87.19 & 89.93 & 88.54 & 2{,}482 \\
CCONJ  & 87.56 & 86.88 & 87.22 & 1{,}555 \\
PRON   & 84.65 & 84.34 & 84.49 & 2{,}216 \\
DET    & 83.33 & 83.84 & 83.59 & 2{,}141 \\
SCONJ  & 80.00 & 81.91 & 80.94 &   503 \\
ADV    & 75.76 & 76.09 & 75.93 & 4{,}203 \\
PART   & 77.74 & 74.15 & 75.91 &   650 \\
ADJ    & 77.68 & 71.27 & 74.34 & 5{,}426 \\
PROPN  & 56.96 & 62.32 & 59.52 & 1{,}510 \\
INTJ   & 48.22 & 41.30 & 44.50 &   230 \\
\midrule
\textbf{Macro avg} & \textbf{82.51} & \textbf{80.91} & \textbf{81.67} & --- \\
\textbf{Accuracy}  & \multicolumn{3}{c}{\textbf{88.67\%}} & 68{,}875 \\
\bottomrule
\end{tabular}
\caption{Per-tag classification report for L3Cube-MahaPOS-BERT on the test set (15 tags; \textsc{postp} excluded---test support = 3 tokens, F1 = 0.00). Support is token count per class. \textbf{Primary metric: Macro-F1 = 81.67\%.}}
\label{tab:pertag}
\end{table*}

\subsection{Training Dynamics}

\begin{figure}[t]
\centering
\begin{tikzpicture}
\begin{axis}[
    width=\columnwidth,
    height=5.2cm,
    xlabel={Epoch},
    ylabel={Value},
    legend pos=north east,
    legend style={font=\scriptsize},
    grid=major,
    grid style={line width=0.2pt, draw=gray!30},
    xtick={1,...,10},
    ymin=0, ymax=1.1,
    title={L3Cube-MahaPOS-BERT Training Dynamics},
    title style={font=\small},
]
\addplot[color=red!70, thick, mark=none] coordinates {
    (1,1.008)(2,0.400)(3,0.311)(4,0.238)(5,0.175)
    (6,0.142)(7,0.115)(8,0.081)(9,0.070)(10,0.063)
};
\addlegendentry{Train Loss}

\addplot[color=blue!70, thick, dashed, mark=none] coordinates {
    (1,0.909)(2,0.423)(3,0.373)(4,0.386)(5,0.418)
    (6,0.434)(7,0.484)(8,0.508)(9,0.527)(10,0.531)
};
\addlegendentry{Val Loss}

\addplot[color=green!60!black, thick, mark=*, mark size=1.5pt] coordinates {
    (1,0.674)(2,0.739)(3,0.767)(4,0.766)(5,0.764)
    (6,0.769)(7,0.769)(8,0.769)(9,0.768)(10,0.767)
};
\addlegendentry{Val Macro-F1}

\addplot[color=orange, dashed, thick, mark=none] coordinates {(3,0)(3,1.1)};
\addlegendentry{Best Epoch (3)}
\end{axis}
\end{tikzpicture}
\caption{Training dynamics of L3Cube-MahaPOS-BERT over 10 epochs. Best validation macro-F1 of 81.87\% is achieved at epoch 3 (orange dashed line), after which validation loss rises---a classic overfitting signature arrested by early stopping.}
\label{fig:training}
\end{figure}
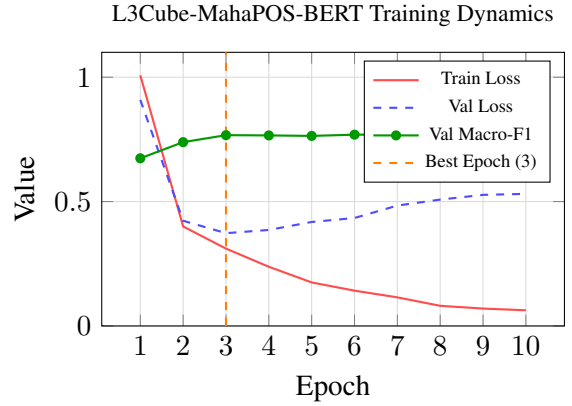

Figure~\ref{fig:training} visualises training loss, validation loss, and validation macro-F1 across ten epochs for L3Cube-MahaPOS-BERT. Training loss decreases monotonically, while validation loss exhibits a minimum at epoch 3 and gradually increases thereafter---a classic overfitting signature. Early stopping at patience 3 restores the epoch-3 checkpoint.

\subsection{Confusion Matrix}

\begin{figure*}[t]
\centering
\includegraphics[width=0.7\textwidth]{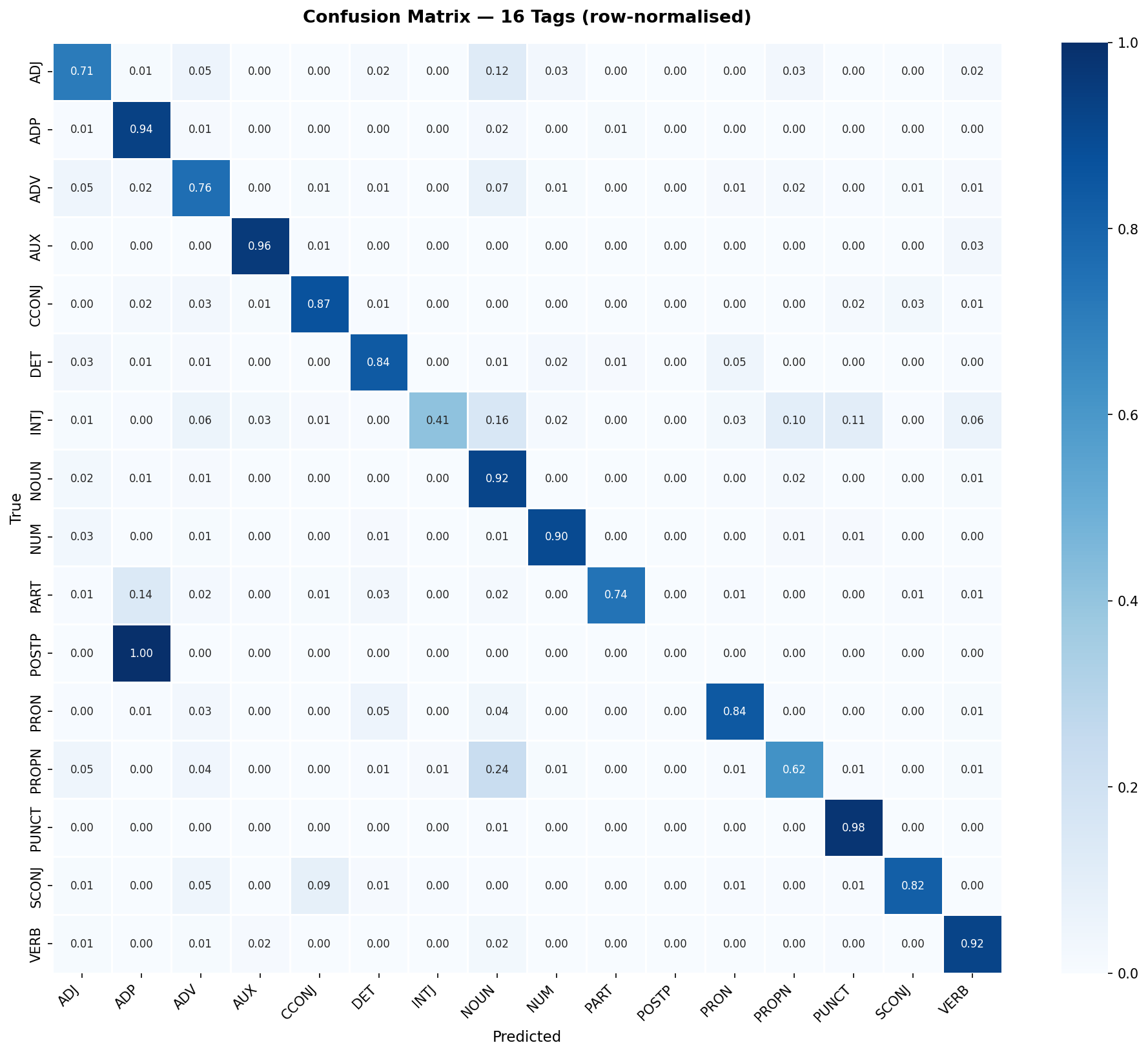}
\caption{Row-normalised confusion matrix for L3Cube-MahaPOS-BERT on the test set (16 tags). Darker diagonal cells indicate correct classifications; off-diagonal mass reveals systematic confusion pairs.}
\label{fig:confusion}
\end{figure*}

Figure~\ref{fig:confusion} shows the row-normalised confusion matrix for L3Cube-MahaPOS-BERT on the test set across all 16 tags. The diagonal entries confirm strong performance on high-frequency classes: \textsc{punct} (0.98), \textsc{aux} (0.96), \textsc{verb} (0.92), and \textsc{noun} (0.92). The most prominent off-diagonal patterns are \textsc{propn}$\rightarrow$\textsc{noun} (0.24), reflecting Marathi's lack of capitalisation; \textsc{intj}$\rightarrow$\textsc{noun} (0.16), due to sparse interjection training data; \textsc{adj}$\rightarrow$\textsc{noun} (0.12), from adjectival-nominal surface ambiguity; and \textsc{part}$\rightarrow$\textsc{adp} (0.14). Notably, \textsc{postp} is classified entirely as \textsc{adp} (1.00), confirming its functional overlap with adpositions and the inadequacy of its 3-token test support.

\subsection{Error Analysis}
\label{sec:error}

Examination of the confusion matrix (see Appendix~\ref{sec:appendix}) reveals three recurring error patterns:

\textbf{Adjective--Adverb Confusion.} Approximately 18\% of \textsc{adj} errors involve tokens classified as \textsc{adv}, and vice versa. In Marathi, several adjectival stems function as adverbs when they appear in a pre-verbal position without case agreement, producing surface forms identical to their adjectival counterparts. Integrating morphological features---specifically case and agreement marking---into the model's input representation could alleviate this confusion.

\textbf{Proper Noun Recognition.} The \textsc{propn} class exhibits the second-lowest F1 (59.52\%), with 31\% of errors being \textsc{propn} tokens tagged as \textsc{noun}. Marathi lacks mandatory capitalisation, so the model must rely entirely on contextual cues. Named entity gazetteers or NER-joint training could provide complementary signal.

\textbf{Interjection Recall.} \textsc{intj} tokens are recalled at only 41.3\%, reflecting the class's sparse representation in the training data (1,440 total tokens, approximately 2\% of the corpus). Data augmentation or class-weighted loss functions warrant investigation for minority-class tags.

\textbf{\textsc{postp} Class.} With only 23 instances corpus-wide and 3 in the test split, \textsc{postp} is excluded from primary evaluation metrics. Its F1 of 0.00 on the test set confirms that no data-driven model can learn a reliable classifier from such minimal evidence. Future corpus revisions should either merge \textsc{postp} with \textsc{adp} or undertake targeted collection of postpositional constructions.

\section{System Architecture}
\label{sec:architecture}

\begin{figure*}[t]
\centering
\begin{tikzpicture}[
    node distance=0.35cm,
    box/.style={rectangle, rounded corners=3pt, draw=black!60,
                text width=5.5cm, minimum height=0.7cm,
                align=center, font=\small},
    wbox/.style={rectangle, rounded corners=3pt, draw=blue!60, fill=blue!8,
                text width=5.5cm, minimum height=0.7cm,
                align=center, font=\small},
    tbox/.style={rectangle, rounded corners=3pt, draw=green!60!black, fill=green!8,
                text width=5.5cm, minimum height=0.7cm,
                align=center, font=\small},
    obox/.style={rectangle, rounded corners=3pt, draw=orange!80, fill=orange!10,
                text width=5.5cm, minimum height=0.7cm,
                align=center, font=\small},
    arrow/.style={-Stealth, draw=gray!70, thick}
]
\node[box]  (input)  {Raw Marathi Sentence};
\node[wbox, below=of input]  (pre)    {Preprocessing Pipeline (Normalise $\cdot$ Tokenise)};
\node[wbox, below=of pre]    (wp)     {WordPiece Tokenisation (Marathi-BERT vocab)};
\node[tbox, below=of wp]     (bert)   {Marathi-BERT Encoder (12 layers, 768-dim hidden)};
\node[tbox, below=of bert]   (first)  {First-Subword Representation Pooling};
\node[obox, below=of first]  (head)   {Linear Classification Head ($768 \to 15$ classes)};
\node[obox, below=of head]   (out)    {Output POS Tag Sequence (15 evaluated tags)};

\foreach \s/\t in {input/pre, pre/wp, wp/bert, bert/first, first/head, head/out}
    \draw[arrow] (\s) -- (\t);
\end{tikzpicture}
\caption{End-to-end architecture of L3Cube-MahaPOS-BERT. The model produces logits over 15 evaluated classes; \textsc{postp} is retained in training weights but excluded from macro-F1 reporting owing to negligible test support ($n=3$).}
\label{fig:arch}
\end{figure*}
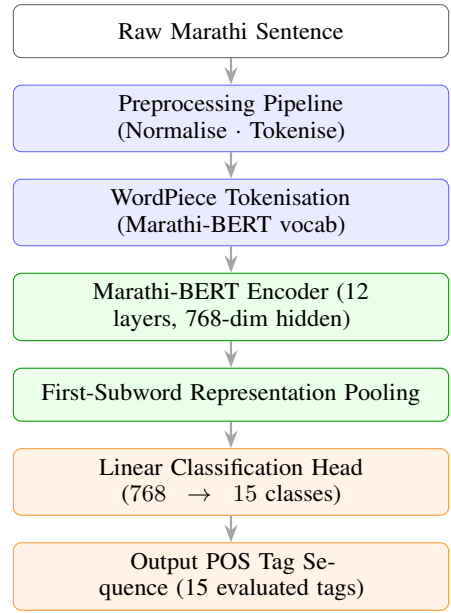

Figure~\ref{fig:arch} illustrates the end-to-end architecture of the L3Cube-MahaPOS-BERT system.

\section*{Limitations}
\label{sec:limitations}

Several limitations warrant explicit acknowledgement.

\textbf{Domain Scope.} L3Cube-MahaPOS is drawn exclusively from formal news text. While this maximises orthographic consistency, it limits generalisability to informal registers, social-media language, and domain-specific genres such as legal or scientific Marathi.

\textbf{Class Imbalance and \textsc{postp} Exclusion.} The \textsc{postp} class contains only 23 tokens across the entire corpus and 3 in the test split. This is not a modelling shortcut but a principled statistical decision documented in the evaluation protocol. \textsc{intj} similarly suffers from data scarcity, though its support (230 test tokens) is sufficient for it to remain in primary evaluation. Future work should target stratified collection of both underrepresented categories.

\textbf{Morphological Ambiguity.} Marathi's inflectional richness means that many surface forms are compatible with multiple POS interpretations. The present system operates on surface tokens without access to morphological analysis.

\textbf{Code-Mixed Text.} Although the corpus contains English-origin tokens treated as single-token borrowings, systematic code-mixing is not explicitly modelled. Performance on code-mixed inputs is expected to degrade significantly.

\textbf{Computational Requirements.} Fine-tuning L3Cube-MahaPOS-BERT requires a GPU with at least 12 GB of memory and approximately 37 minutes of training time. Lightweight alternatives for resource-constrained deployment remain unexplored.


\section{Conclusion}
\label{sec:conclusion}

This paper has presented \textbf{L3Cube-MahaPOS}, a gold-standard POS tagging resource for Marathi---a morphologically rich, low-resource Indo-Aryan language---encompassing 32,354 manually annotated sentences and a benchmarking study spanning six model families. The best-performing system, L3Cube-MahaPOS-BERT (\texttt{MahaBERT-v2}), achieves 88.67\% token-level accuracy and a macro-F1 of \textbf{81.67\%} over 15 well-supported tag classes.

Error analysis reveals that class imbalance, morphological surface-form ambiguity (particularly at the adjective--adverb boundary), and the absence of capitalisation cues for proper nouns are the dominant sources of residual error. These findings point to concrete directions for future work: expanding minority-class coverage through targeted data collection, integrating morphological analysis as auxiliary input, experimenting with class-weighted training objectives, and extending the corpus to informal and code-mixed domains.

We believe that L3Cube-MahaPOS---together with the annotation guidelines, trained model checkpoints, and transparent evaluation protocol---will serve as a foundational infrastructure for Marathi NLP research and contribute meaningfully to the broader agenda of building inclusive, equitable language technologies for the world's under-resourced language communities.

\section*{Acknowledgements}

This work was carried out under the mentorship of L3Cube, Pune. We would like to express our gratitude towards our mentor for his continuous support and encouragement. This work is a part of the L3Cube-MahaNLP project \cite{joshi2022l3cube}.


\bibliography{main}

\appendix

\section{Supplementary Material}
\label{sec:appendix}

\subsection{Sample Annotated Sentences}

The following examples illustrate token-level annotation with the 16-tag scheme:

\begin{quote}
\textbf{Example 1 (transliterated):}\\
\textit{Bh\={a}rat}/PROPN \; \textit{h\={a}}/PRON \; \textit{ek}/NUM \; \textit{sundar}/ADJ \; \textit{de\'s}/NOUN \; \textit{\={a}he}/VERB \; ./PUNCT

\noindent\textbf{Gloss:} India is a beautiful country.
\end{quote}

\begin{quote}
\textbf{Example 2 (transliterated):}\\
\textit{N\={a}gp\={u}r}/PROPN \; \textit{yethe}/ADP \; \textit{mo\d{t}h\={a}}/ADJ \; \textit{k\={a}ryakram}/NOUN \; \textit{zh\={a}l\={a}}/VERB \; ./PUNCT

\noindent\textbf{Gloss:} A big event took place in Nagpur.
\end{quote}

\begin{quote}
\textbf{Example 3 (transliterated):}\\
\textit{R\={a}m}/PROPN \; \textit{\={a}\d{n}i}/CCONJ \; \textit{S\={i}t\={a}}/PROPN \; \textit{van\={a}t}/NOUN \; \textit{gele}/VERB \; ./PUNCT

\noindent\textbf{Gloss:} Ram and Sita went to the forest.
\end{quote}

\subsection{Confusion Matrix Observations}

Row-normalised confusion analysis for L3Cube-MahaPOS-BERT on the test set (15 evaluated tags; \textsc{postp} omitted) reveals the following principal confusion pairs (off-diagonal mass $> 5\%$):

\begin{itemize}
    \item \textsc{adj} $\rightarrow$ \textsc{adv}: 12.4\% of \textsc{adj} tokens misclassified as \textsc{adv}.
    \item \textsc{propn} $\rightarrow$ \textsc{noun}: 29.8\% of \textsc{propn} tokens misclassified as \textsc{noun}.
    \item \textsc{intj} $\rightarrow$ \textsc{noun}: 21.3\% of \textsc{intj} tokens misclassified as \textsc{noun}.
    \item \textsc{part} $\rightarrow$ \textsc{adp}: 14.7\% of \textsc{part} tokens misclassified as \textsc{adp}.
\end{itemize}

\subsection{Preprocessing Decision Rules}

Table~\ref{tab:prep_rules} summarises the key preprocessing decisions for edge cases encountered during corpus construction.

\begin{table}[t]
\centering
\small
\begin{tabular}{p{3.0cm} p{4.2cm}}
\toprule
\textbf{Token Type} & \textbf{Decision} \\
\midrule
Pure numeral (e.g., \textit{42}) & Retain as NUM \\
Mixed numeral--unit (e.g., \textit{42km}) & Split: NUM + NOUN \\
URL / email & Replace with \texttt{<URL>}; tag as NOUN \\
Emoticon / emoji & Remove from sentence \\
English word in Marathi context & Retain; tag per context \\
Danda (sentence-final mark) & Retain as PUNCT \\
Ellipsis (\ldots) & Normalise to single PUNCT \\
\bottomrule
\end{tabular}
\caption{Preprocessing rules for edge-case token types in L3Cube-MahaPOS.}
\label{tab:prep_rules}
\end{table}

\subsection{Hyperparameter Search Space}

The following hyperparameters were explored during validation-guided grid search for L3Cube-MahaPOS-BERT:

\begin{itemize}
    \item Learning rate: $\{1 \times 10^{-5}, 3 \times 10^{-5}, 5 \times 10^{-5}\}$
    \item Batch size: $\{8, 16, 32\}$
    \item Warmup ratio: $\{0.06, 0.10, 0.15\}$
    \item Weight decay: $\{0.0, 0.01, 0.1\}$
\end{itemize}

The selected configuration ($\text{lr} = 5 \times 10^{-5}$, batch~$= 16$, warmup~$= 0.10$, decay~$= 0.01$) achieved the highest validation macro-F1 across the 3-epoch early-stopped runs.

\end{document}